\documentclass[runningheads]{llncs}
\usepackage{graphicx}
\usepackage{amsmath}
\usepackage{url}
\usepackage{color}

\begin{document}
\title{Identifying centres of interest in paintings \\ using alignment and edge detection. \\ \large{Case studies on works by Luc Tuymans} \thanks{This work was made possible thanks to a 'scientist in residence 
of an artist studio' grant to Luc Steels and Studio Luc Tuymans. The residency was funded by the 
European Commission's S+T+ARTS programme, set up by DG CONNECT. It was organized by BOZAR, a Belgian art centre located in Brussels, and GLUON, a Brussels organisation facilitating art-science interactions. Additional funding for this research has come from the H2020 EU project MUHAI on Meaning and Understanding in AI.}}
\titlerunning{Detecting centres of interest in paintings}
%
\author{Sinem Aslan \inst{1} \orcidID{0000-0003-0068-6551} \and Luc Steels \inst{2}\orcidID{0000-0001-9134-3663}}
\authorrunning{S. Aslan and L. Steels}
%
\institute{Ege University \\ International Computer Institute, Izmir, Turkey \\ \email{siinem@gmail.com}\\
\url{http://akademik.ube.ege.edu.tr/~aslan/} \and Catalan Institut for Advanced Studies (ICREA)\\ Institute for Evolutionary Biology (UPF-CSIC) Barcelona \\ \email{steels@arti.vub.ac.be}\\
\url{https://www.icrea.cat/Web/ScientificStaff/Steels-Luc-539}}
\maketitle              
\begin{abstract}
What is the creative process through which an artist goes from an original image
to a painting? Can we examine this process using techniques from computer 
vision and pattern recognition? Here we set the first preliminary steps 
to algorithmically deconstruct some of the transformations that an artist applies
to an original image in order to establish centres of interest, which 
are focal areas of a painting that carry meaning. We introduce 
a comparative methodology that first cuts out the minimal segment from the original 
image on which the painting is based, then aligns the painting with this source, investigates
micro-differences to identify centres of interest
and attempts to understand their role. In this paper we focus exclusively
on micro-differences with respect to edges. We believe that research into where and how 
artists create centres of interest in paintings is 
valuable for curators, art historians, viewers, and art educators, and might 
even help artists to understand and refine their own artistic method.
\end{abstract}
\setcounter{footnote}{0}
\section{Introduction}

There is already a significant body of research investigating artistic paintings
using techniques from pattern recognition, computer vision and AI. For example, 
some researchers have used deep learning to extract patterns from a series of paintings
in order to obtain a statistical model of the painter's style and then create new images 
in the same style \cite{SemmoIsenbergDollner:2017}. Others use 
neural networks for classifying paintings \cite{Cetinic:2018} or semantic web technology 
for organizing large collections based on rich ontologies so that these collections
can be searched in more powerful ways \cite{deBoer:2013}. In addition, many projects 
have been using image processing to assist in 
conservation work, artist identification, detection of forgery, and many other 
applications \cite{Cornelis:2015}. 
This paper focuses on a new, complementary aspect of art investigation, which 
attempts to {\it understand the artistic creation process through AI modeling}.
\\
\indent We start from the hypothesis that an artist tries to stimulate
a process of narrative creation in the viewer and uses the expressive means available in 
the chosen medium. In the case of painting, this includes objects and figures being depicted, 
lines, colors, contrast, composition, 
blurring of background, etc. The artist is in a sense 
engaged in a form of cognitive engineering \cite{Dewey:2018}, manipulating the mental 
processes of viewers by shaping their sensory experiences and memory recalls.   
The creation of an artistic painting is therefore more than 
the application of a particular style to an existing image, as done in deep
generative adversarial neural network experiments \cite{SemmoIsenbergDollner:2017}. 
A style can certainly 
be recognized by current AI technology and even replicated, but the creation of an artwork is 
more than achieving formal appearances in a particular style. It is about 
choosing a pertinent subject, transforming images into more powerful ones, and evoking  
a web of meanings in the viewer. 
\\
\indent Consequently, the interpretation 
of an artwork has a strong similarity with the {\it understanding} of a text and the 
creation of an artwork is similar to the {\it production} of a text, with the important 
difference that many of the meanings expressed or invoked by paintings or 
works in other media, like music or dance, are pre-verbal. They concern 
conceptualizations, emotions, moral values and perspectives that are not so easy to 
put into words. These meanings resonate with viewers 
at a subconscious level and, partly for this reason, they may evoke a stronger reaction 
from the viewer than obtainable from a purely rational text. 
Art works activate the non-rational, non-cognitive areas of the brain, such as the emotion 
system \cite{Ledoux:1996} or the social brain \cite{Vilarroya:2007}. 
\\
\indent A work of art is complex and rich because it evokes the different levels of meaning
(formal, factual, expressional, cultural, intrinsic) 
at the same time and the meanings at each level intimately interact with others \cite{Panofsky:1939}. 
Moreover which set of meanings resonates with one viewer is usually not the same as 
those resonating with another viewer or with those originally felt or intended by the 
artist, simply because everybody has their own episodic and 
semantic memory, their own prior experiences 
of artworks, their own social context and psychological state when viewing, experiencing
and interpreting an artwork. There is therefore no objectively `correct' set of meanings 
and it would be futile to develop an AI system that would extract such a set. 
However, this paper will try to show that we can nevertheless 
study {\it artistic methods}, i.e. the kind of transformations 
an artist performs on a real scene or a found image in order to induce meanings in the viewer.
\\
\indent An important part of the artistic painting method consists in the introduction of visual 
{\it centres of interest}, attention getters which guide meaning invocation. 
They are instantiated partly by what is being represented, for example, an iconic 
image of a well known person which is appropriated by the artist, such as
the well-known image of Marilyn Monroe used by Andy Warhol. Partly there are 
objects or parts of objects where the human vision system naturally pays 
attention to, for example the eyes of a face. And in addition, artists often 
introduce deviations from what is expected, for example they deform the nose on a face, 
or they use devices such as contrast, bright 
colors, sharp edges, blurring of background, flattening out of 3d effect, etc.
\\
\indent There is often a first most salient centre of attention, called the primary 
focal point, which pulls the viewer into the painting. For example, in the case of 
Figure \ref{fig:K-painting} (left), the focal point is clearly the left eye and to a second 
degree the lips. The primary focal point introduces an organizing perspective
from which the gaze of the viewer starts its exploration. 
Then there are additional focal points, as the viewers' eye gaze glances 
over the different parts of the painting. 
\\
\indent In earlier work in collaboration with Bj\"{o}rn 
Wahle \cite{SteelsWahle:2020}, we already used various computer vision methods to investigate 
how saliency identifies the primary focal point. The first set of 
saliency detection methods we tried on the Tuymans paintings
were low-level, in the sense that they estimate saliency based on general 
statististical properties of the visual image \cite{SteelsWahle:2020}. 
In this paper we introduce a complementary new and quite different methodology to study which 
areas in a painting are interest centres. The methodology is 
based on comparing an original source image (for example a photograph)
with the painting based on it and deconstructing the interventions 
the painter undertook to transform one into the other. Many contemporary painters expressly use 
existing images from popular culture to make contact with their audience and thus comment on 
contemporary media culture. Our comparative methodology contains four steps: 
\newline
\indent (i) Locate the original image, either directly from the painter or 
searching information resources like 
images available on the world wide web or in image repositories using reverse image search. 
\newline
\indent (ii) Align the painting with the original image. This 
involves finding the minimal set of {\it macro-operations} performed by the artist, such as 
cutting, rotating, or scaling so that the contents of the painting are optimally aligned with the 
corresponding segment in the original image. 
\newline
\indent (iii) Zoom in on regions where there have been {\it micro-transformations} focusing on 
specific aspects of the painting, for example changes in 
shape, blurring, changes in illumination, edges differences, color changes, etc. The 
difference map between original image and painting for each of these aspects provides us 
with hypotheses what areas may be centres of interest and triggers inquiries in their possible 
meaning. 
\newline
\indent (iv) Importantly we then want to understand the meanings of these macro- and 
micro-transformations. This cannot be done at the moment with AI but introspection and  
conversations with the artist will help us to move in this direction in the future. In any 
case we have found that the suggestions coming out of the previous steps are very 
illuminating because they tell us where to look. In any case, they have enriched the experience of 
paintings by the present authors. 

\section{Case studies}

The present paper reports the first results from applying our comparative methodology using
paintings by the contemporary Flemish painter Luc Tuymans. 
Working with a living artist makes it possible to validate our 
hypotheses about what the centres of interest are in a particular painting
and they can tell us whether AI techniques
have yielded anything worthwhile, not only for viewers, curators or art historians but 
also for those creating artworks themselves. 

Luc Tuymans is considered
to be one of the most important contemporary painters at the moment \cite{Loock:1996}.
He has done solo exhibitions
at some of the most prestigious and influential
art centres in the world such as the MOMA in New York, 
the Palazzo Grassi in Venice, the MCA Museum of Contemporary Art in Chicago, BOZAR in Brussels, 
the St\"{a}del Museum in Frankfurt, the National Art Museum of Beijing, etc. 
We have been fortunate because we have a direct and recurrent contact with this painter 
and have access to the relevant parts of his digital 
archives. Moreover Luc Tuymans is very articulate in describing his own artistic method 
as well as the methods used by other painters \cite{Tuymans:2018}. 

\begin{figure}[!h]
  \vspace{-0.2cm}
  \centering
  \includegraphics[width=0.5\linewidth]{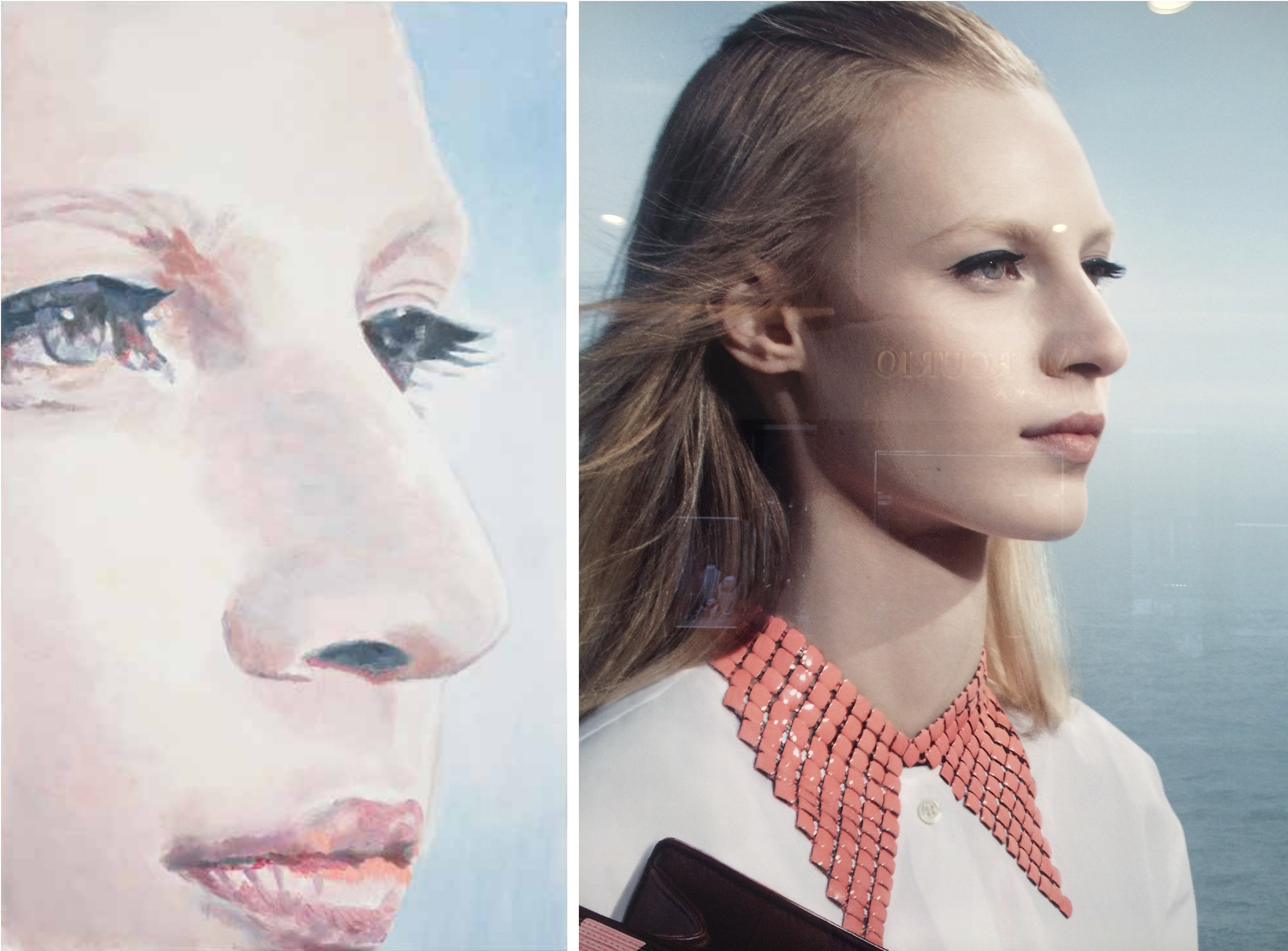}
  \caption{Left: `K.' by Luc Tuymans, 2017, oil on canvas. 135 $\times$ 
80,2 cm. Andrew Xue Collection, Singapore. Right: Source image as provided by the artist.}
  \label{fig:K-painting}
  \vspace{-0.1cm}
\end{figure}

We have constructed a general interactive 
pipeline of pattern recognition, computer vision and AI algorithms 
to apply our methodology and processed several paintings by Luc Tuymans. Here we only 
focus on those parts of the pipeline that involve alignment and edge detection. 
The paintings we have studied in detail so far have come from the 
solo exhibition of Luc Tuymans at the Palazzo Grassi in Venice (2019-2020) but we have 
also done experiments on all the paintings Luc Tuymans documented in his 
Catalogue Raison\'{e}\cite{Meyer-Hermann:2019}. 
Given space limitations, we focus in this paper mainly on one specific painting, 
entitled K., shown in Figure \ref{fig:K-painting} (left) where the focus on edge difference maps 
has been fruitful. But we also show a second
example, entitled Secrets, shown in Figure \ref{fig:secrets-painting-original-alignment}, 
where a focus on edge differences has yielded less results, mainly because the 
changes at the edge level are too numerous - hence we get too many possible centres of interest. 

\section{STEP I. Finding the original image}
\vspace{-0.15cm}
Due to direct contacts with the artist, we have had access to several original sources 
for the Palazzo Grassi paintings, including the one 
for K. (see Figure \ref{fig:K-painting}, right). But we also wondered whether, with the 
massive availability of images on the web and the sophistication of current image recognition 
technology, it was possible to find the original source image using {\it reverse image search}
with the painting as a key. 
\\
\indent Interestingly, a reverse image web search with the painting K. did not deliver anything 
close to a possible source image for K., even though the source was also available 
on the web (see below). 
It appears that there are properties of artistic paintings
which make the use of reverse image search algorithms difficult, whereas
a human observer immediately recognizes that the painting on the left of 
Figure \ref{fig:K-painting} depicts the woman shown on the right. 
The issue of domain adaptation has become a recent hot topic in computer
vision research but we have not applied these new methods yet \cite{wilson2020survey}.

\begin{figure}[!h]
  \vspace{-0.2cm}
  \centering
  \includegraphics[width=0.5\linewidth]{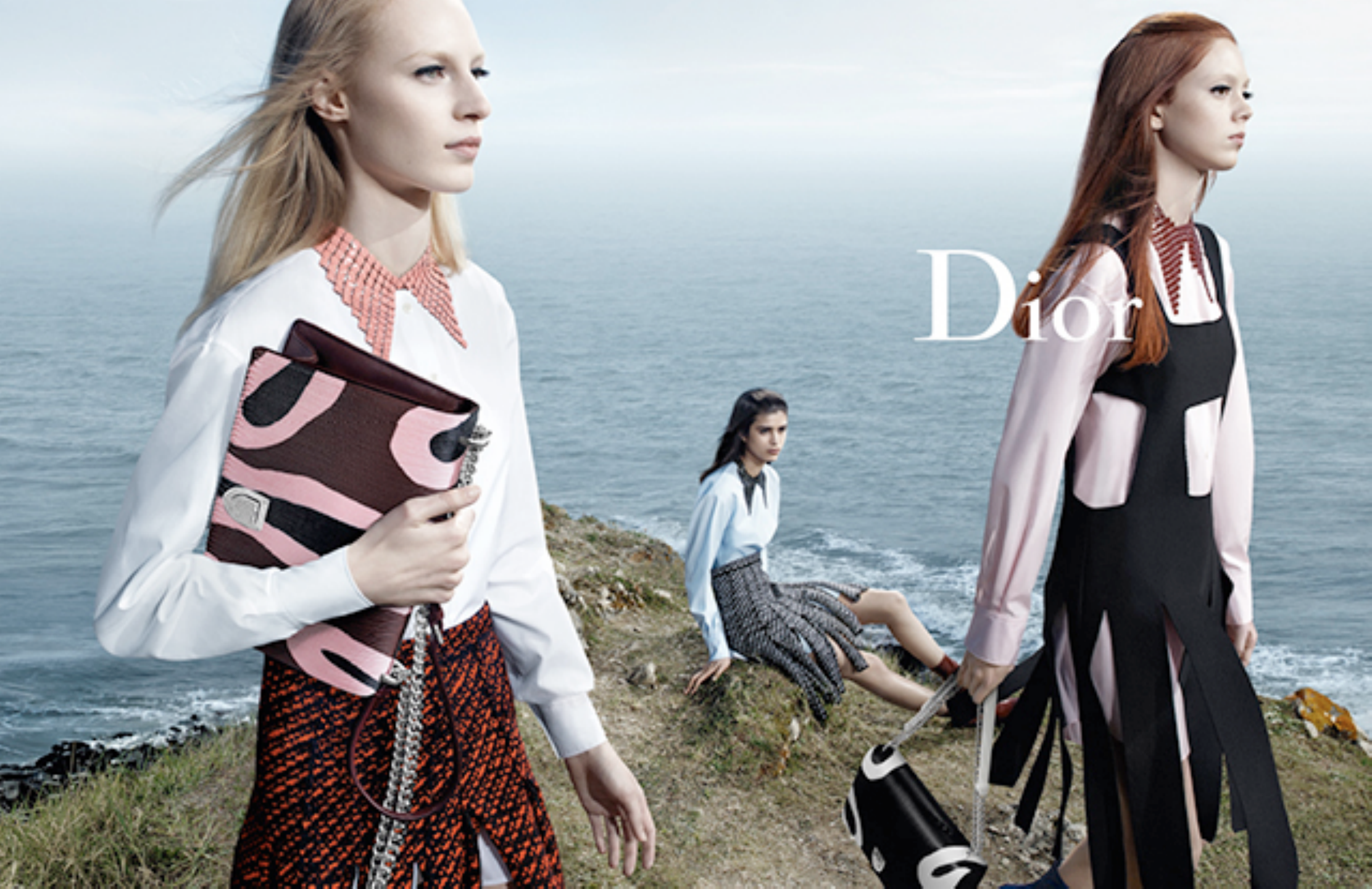}
  \caption{Image from the Dior Autumn-Winter 2015 campaign with clothes designed by Raf Simons 
and photography by Willy Vanderperre. The fashion model to the left, the basis of K., is Julia Nobis.}
  \label{fig:dior}
  \vspace{-0.1cm}
\end{figure}

On the other hand, once an original source image is available, reverse image search systems,  
like the one provided by Google or Bing, {\it are} 
able to locate this original and its variants, even if the provided original 
image is only a segment within a larger image or parts of the original image
are not included. Thus using the original image in Figure \ref{fig:K-painting} (right), 
we found the actual context of 
the original image. It turned out to come from an advertising campaign by Dior (see 
Figure \ref{fig:dior}), showing a clothing line designed by fashion designer Raf Simons and
photographed by Willy Vanderperre, both alumni of the Antwerp art academy where Tuymans 
studied as well. The scene has been set up on the Normandy coast in France. 
Notice that in the advertisement, the top hair of K. is cut 
out which is not the case in the original image for K. provided by the painter. 

Finding the original image of a painting is of interest from an art-history point of view and it 
provides important cues as to the factual and cultural meaning of the painting. This kind of 
fashion advertisement imagery - not necessarily this particular image - is familiar 
to everybody through magazines and posters and it reflects contemporary culture. 
Luc Tuymans expropriates the esthetic elements and the staging of the fashion models but 
at the same time removes completely the original context in order to create 
a timeless image. 

\section{STEP II. Aligning the painting and the original} 
\vspace{-0.15cm}
The next step in our methodology is to align the original image with the painting. This is a very non-trivial image processing problem. For this paper, we did a manual selection of a possible candidate
image, cutting out a segment that left out as much as possible of the material that was 
not in the painting, because we found that the vision algorithms we tried do not work well when there 
was too much additional extra material in the source image. 

Given a good candidate, the alignment problem becomes similar to the so called 
{\it image registration task} \cite{goshtasby2005} which is widely used in medical imaging, automated manufacturing, satellite navigation, and many other application fields for comparing or integrating data acquired from different sensors, at different times, or with different depth or viewpoints. Image registration algorithms fuse multimodal information or detect changes on such images. Given a set of two images, the one that will be transformed is called the {\it moving image}, and the one that is left unchanged, is called the {\it reference image}.

In the present investigation, we view the original image (the photograph) as the moving image and try to align it to the painting by progressively transforming the original image. The painting is therefore the reference image in image registration terminology. This is the perspective of the painter because the painter transforms the original image into a painting. Conversely, we can start from the painting, which now becomes the moving image, and try to align it to the original image, which now becomes the 
reference image, by transforming the painting to look similar to the photograph. 
This is in a sense the perspective from the viewer who sees the painting and then tries to align it with an image stored in his or her episodic memory. Both approaches yield interesting results, but in this paper we focus mainly on the painter's perspective only.

Image registration has been discussed intensely in the computer vision literature and many algorithms exist, primarily feature-based or intensity-based methods \cite{goshtasby2005}. Feature-based methods aim to find a correspondence between image features such as interest points, lines, and contours, while intensity-based methods aim to align pixel patterns via correlation or similarity metrics. In this paper, we will rely only on the multi-modal intensity-based registration method, 
implemented using the Matlab Registration 
Estimator App with its default parameters.\footnote{\url{https://www.mathworks.com/help/images/register-images-using-the-registration-estimator-app.html}}

We also experimented with the feature-based approach trying to detect the candidate features on source and target images by various well-known algorithms, in particular, the {\it Maximally stable extremal region method}
(MSER)\cite{Donoser:2006} and the {\it Speeded up Robust Features Method} (SURF) \cite{Bay:2008}. 
However, we found that the output of the feature-based approaches performed poorly
for the alignment task, possibly due to the significant gap between source and target images, as one is a photograph and the other a painting. Using features from these classical methods leads to the detection of too many low quality interest points to yield effective alignment. 
\begin{figure}[!h]
  \vspace{-0.2cm}
  \centering
  \includegraphics[width=0.8\linewidth]{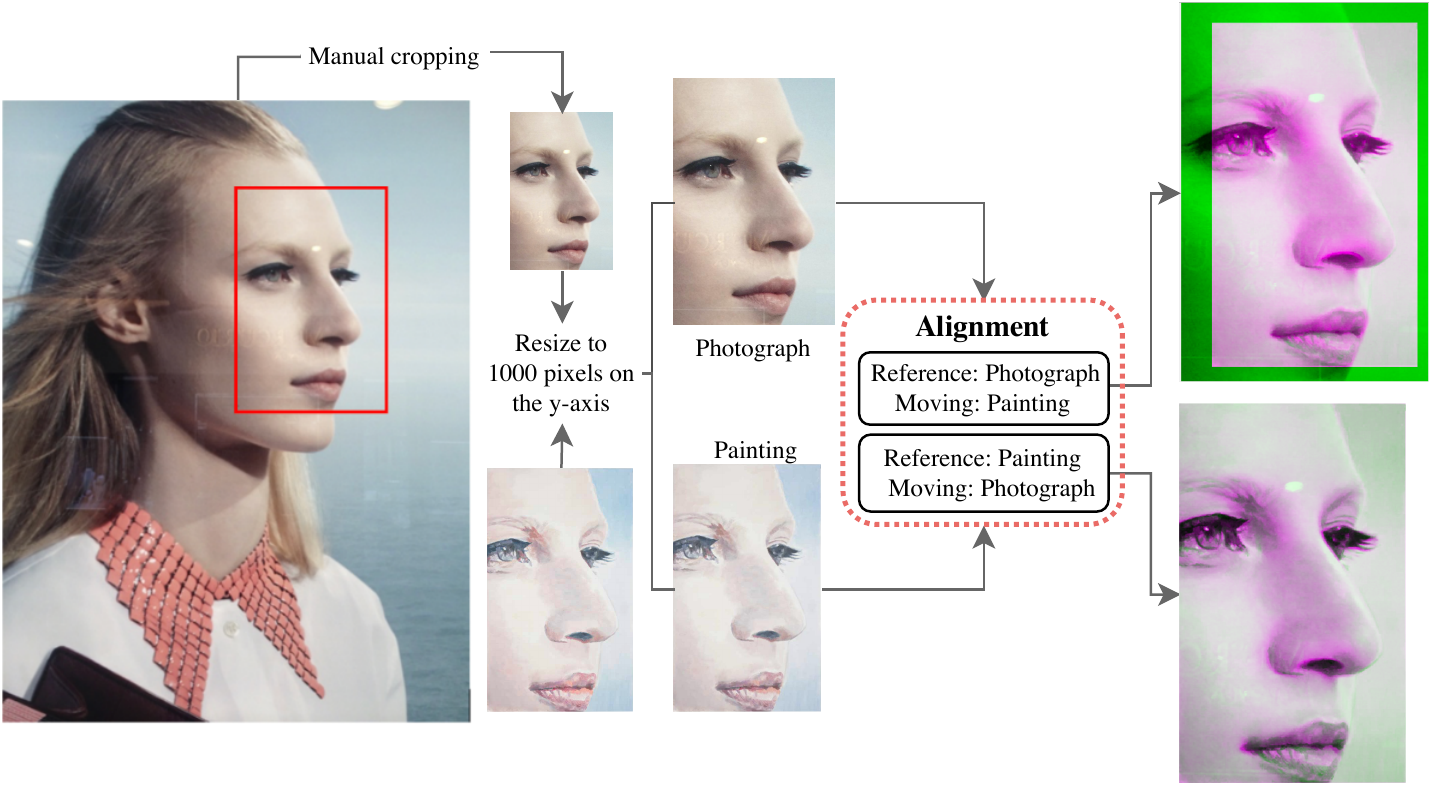}
  \caption{Flow diagram for image alignment of K. 
The top show the `viewer's perspective', using the painting as moving image and the photograph
as reference image, and at the bottom, we show the `painter's perspective' with
the photography as moving image and the painting as reference image. To show
the overlay, the painting is rendered using a purple color and the original 
photograph using green.}
  \label{fig:flow_registration}
  \vspace{-0.1cm}
\end{figure}
\\
\indent We operationalized alignment using a best-first search algorithm, as used for example in \cite{styner2000parametric}. The Mattes mutual information metric \cite{rahunathan2005image} is used to compute the similarity between source and target. It is shown in \cite{rahunathan2005image} that this metric leads to better alignment than the Mean Squares metric when the goal is to align images from different modalities using rigid transformations, which is the case here. 
The algorithm searches for a set of transformation parameters that produce the best possible alignment result. Given a matrix of transformation parameters M, called the parent, a number of variations $M_1$, ... $M_n$, called the children, are generated, at first using 
aggressive perturbations. The perturbations include shape-preserving transformations like 
\textit{rotation}, \textit{translation}, \textit{isotropic scaling}, and \textit{reflection}. 
If a child $M_i$'s parameters yield a better alignment, then it 
becomes the new parent on the next iteration. But if a parent still yields a better result, 
it remains as a parent and new children are computed with less aggressive changes to the 
parent's matrix.  

\section{STEP III. Micro-transformations} 
\begin{figure}[!h]
  \vspace{-0.2cm}
  \centering
  \includegraphics[width=0.7\linewidth]{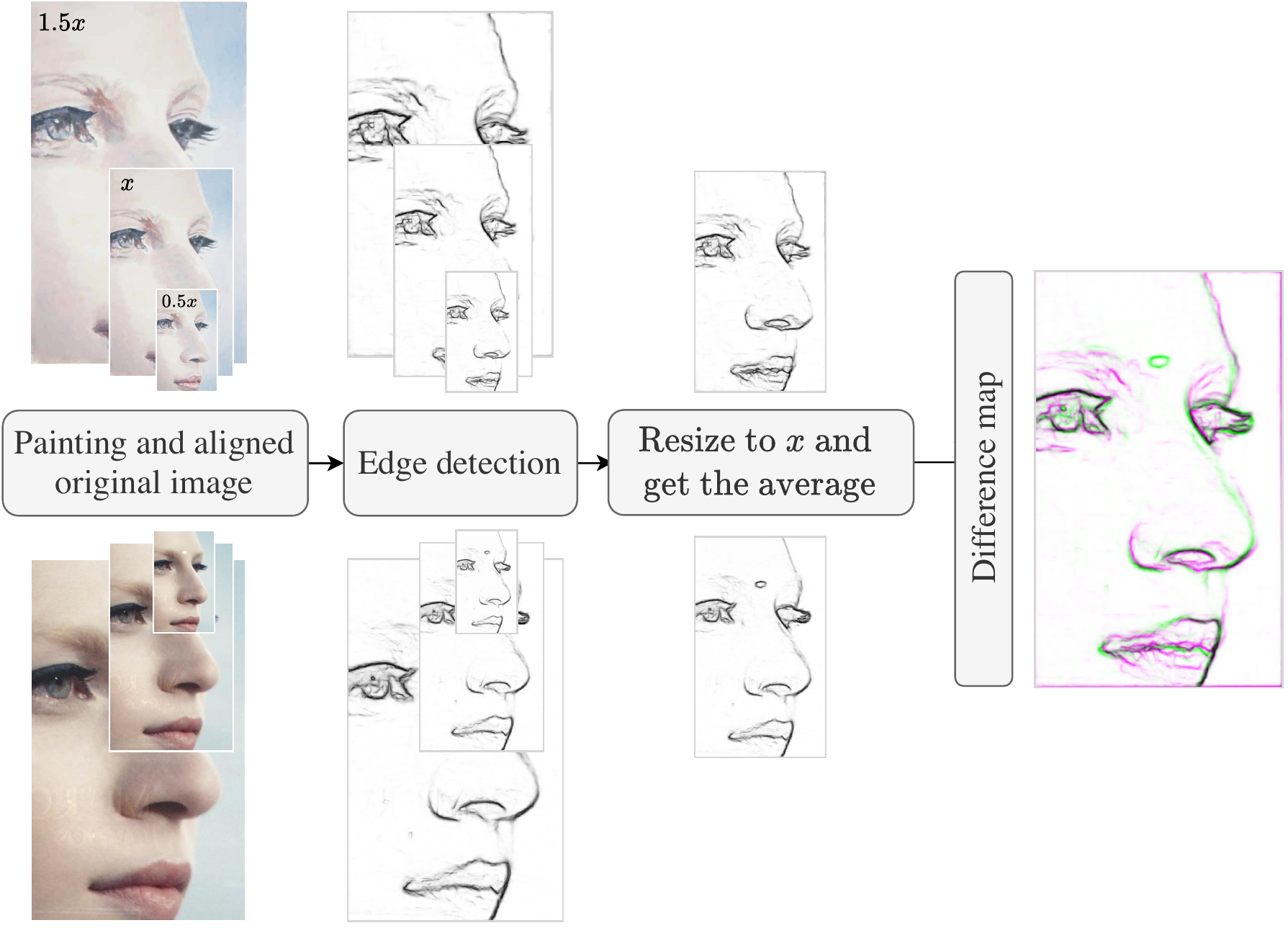}
  \caption{Flow diagram for edge detection. 
It shows the edge maps for K. on top for the painting and below for the original 
image at different scales based on the TIN method. 
The difference map is shown with purple for the photograph, 
green for the painting and black when edges overlap.}
  \label{fig:flow_edges}
  \vspace{-0.1cm}
\end{figure}

\vspace{-0.15cm}Once we have adequately aligned the original image with the painting, it 
becomes possible to inquire about the micro-transformations that the painter has introduced and 
their function. These variations have happened for different visual aspects, e.g. color, contrast,
figure orientation, contours, etc. and so we need to first isolate these aspects from the image. 
In the rest of this paper we only look at edges, which means that we investigate which 
additional edges or edge variations the painter has introduced, 
The flow diagram of this pipeline is illustrated in Fig. \ref{fig:flow_edges}. 
We have experimented with two algorithms, a traditional edge detection method, namely 
the {\it Sobel Isotropic $3\times3$ gradient operator} (SOBEL), and a deep neural network 
known as the {\it Traditional Inspired Network} (TIN) recently introduced by Wibisono 
and Hang \cite{Wibisono2020TIN}. The SOBEL edge detection method
has been a very popular and widely used algorithm in image processing since 1968
\cite{sobel1968isotropic}. It is based on derivation of a simple and computationally efficient gradient operator. In order to calculate the approximations of derivatives for horizontal and vertical intensity changes, the gray-scaled input image is convolved with two $3 \times 3$ kernels as follows :
\[
G_x =\begin{bmatrix}
    +1  & \quad   0  &  \quad  -1 \\
    +2  &  \quad  0  &  \quad  -2  \\
    +1  &  \quad  0  &  \quad  -1    
\end{bmatrix} \ast I
\quad \quad 
G_y =\begin{bmatrix}
   +1 &  \quad  +2  & \quad  +1 \\
     0  &  \quad    0  & \quad     0  \\
   -1  &  \quad - 2  & \quad   -1    
\end{bmatrix} \ast I
\]
where $I$ denotes the input image in grayscale, $G_x$ and $G_y$ are two images which at each point contain the vertical and horizontal derivative approximations. 
Then, the resulting gradient approximations are combined to have the gradient magnitude at each point in the image, using $G = \sqrt{G_x^2+G_y^2}$. Subsequent images showing the results of the Sobel Method, display this gradient magnitude $G$. 

Deep neural network frameworks have originally been designed for high-level computer vision tasks such as object recognition or scene understanding through semantic segmentation. Edge detection is a more local and simpler task so that a lightweight deep learning network can provide high quality edges with reduced computational complexity. This approach has been adopted by the \textit{Traditional Method Inspired Network} (\textit{TIN}) proposed very recently by Wibisono and Hang \cite{Wibisono2020TIN} with reported state of the art accuracy performances on the BSDS500 test set. The framework is composed of three modules, i.e., \textit{Feature Extractor}, \textit{Enrichment}, and \textit{Summarizer}, which roughly correspond to gradient, low pass filter, and pixel connection in the traditional edge detection schemes. 
The pre-training of the TIN method was performed on three datasets of natural images BSDS500, 
In our experiment, we used the published code\footnote{\url{https://github.com/jannctu/TIN}} by the authors and we used their pretrained model on the aforementioned datasets for the proposed architecture TIN2, since higher performances were reported by TIN2 compared to TIN1 in \cite{Wibisono2020TIN}. 
\begin{figure}[!h]
  \vspace{-0.2cm}
  \centering
  \includegraphics[width=0.25\linewidth]{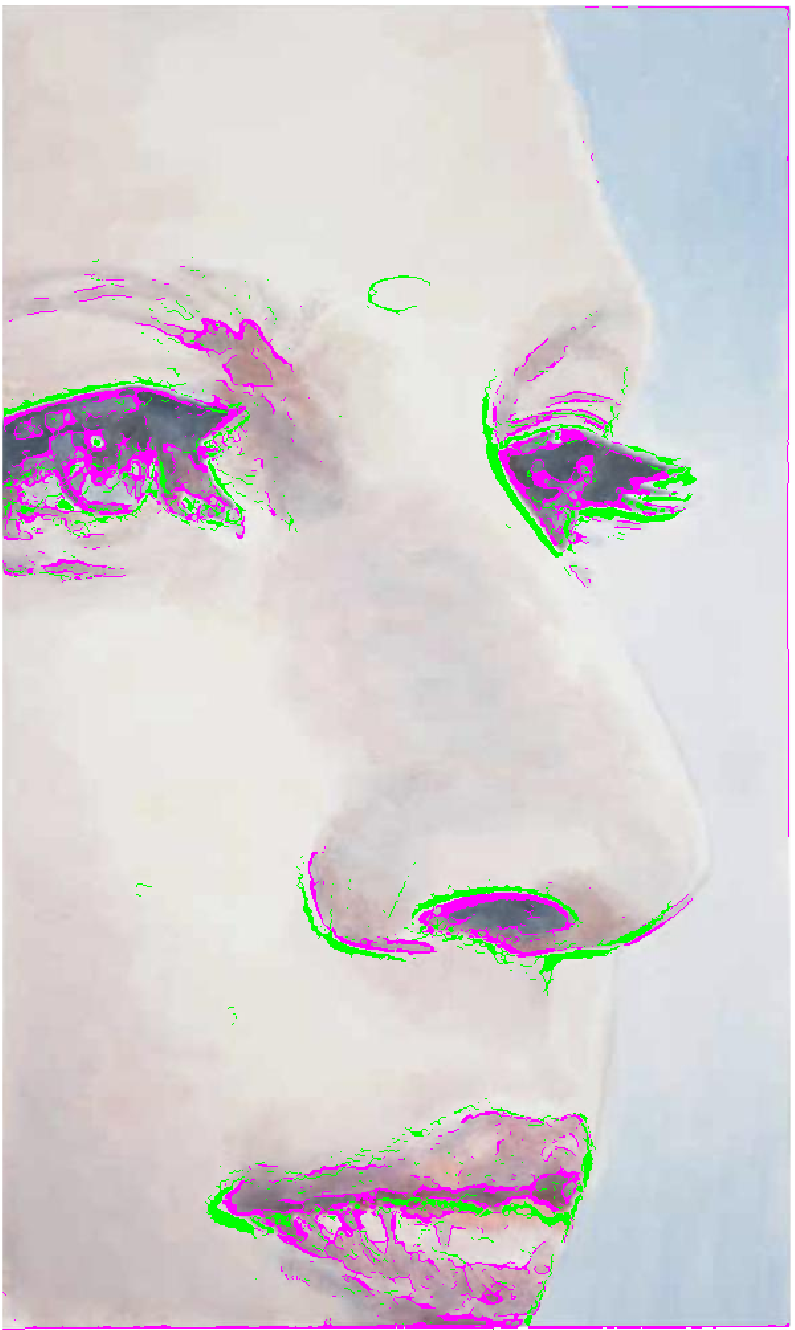}
  \includegraphics[width=0.25\linewidth]{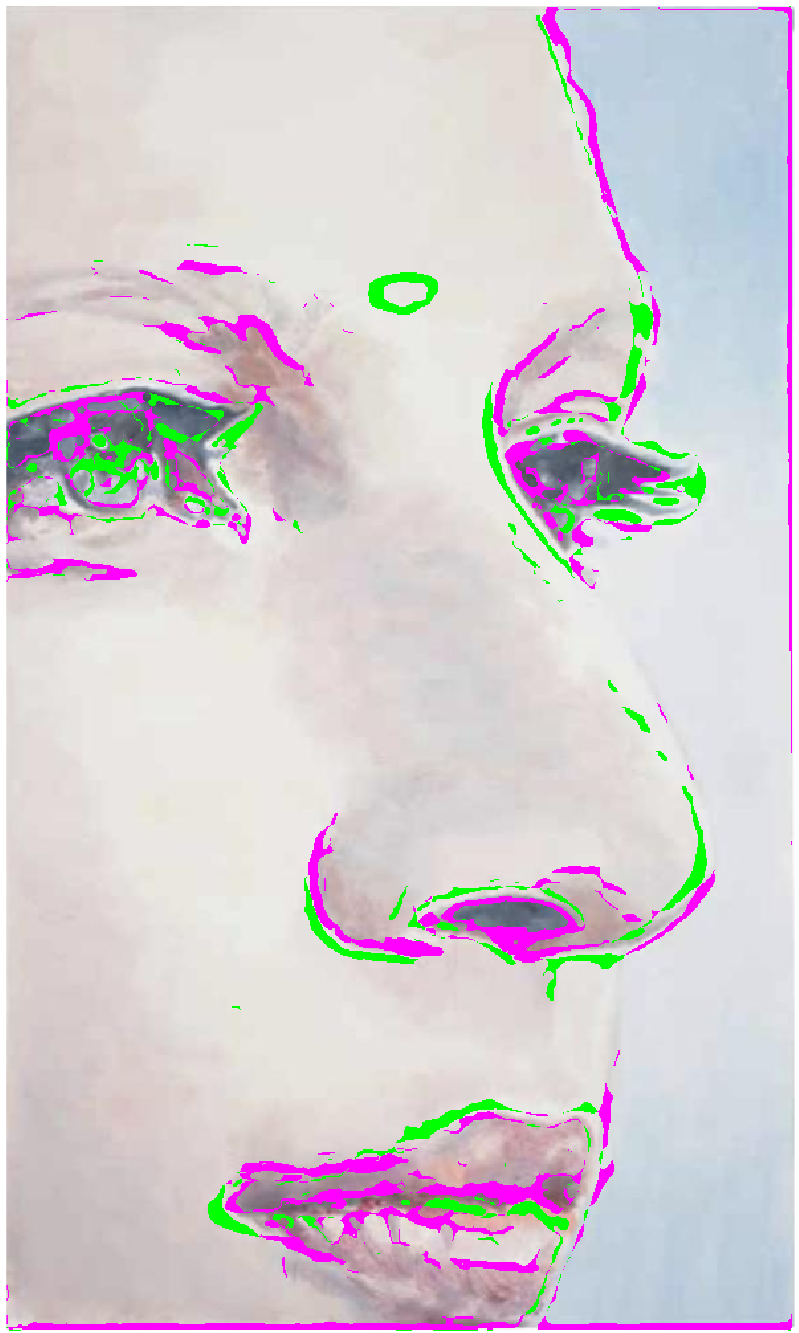}
  \caption{Comparison of two edge detection methods, taking the painter's perspective
(going from photograph to painting). The painting edges are in green and the photograph
edges in purple. Left: the difference map for the SOBEL method. Middle: 
the difference map for the TIN method. We see that the TIN method provides clearer edges
showing the variations introduced by the painter in a clearer way.}
  \label{fig:compare-edge-detection}
  \vspace{-0.1cm}
\end{figure}
To make a fair comparative analysis between the SOBEL and TIN methods, we applied the same post-processing operation, namely {\it Non-Maximum Suppression} (NMS), to both edge-detection methods, i.e. we computed edge maps at different scales, $1.5x$, $1x$, $0.5x$, for both the origin and target, took the average of these maps, and resized the edge maps to the original image size, i.e., $x=1000$. As mentioned in \cite{Wibisono2020TIN}, this procedure provides 
better-located edges. 

Finally, we compute the edge difference maps for further analysis. Figure \ref{fig:compare-edge-detection} shows 
the outcome after thresholding in order 
to binarize the edge map so that the significant edges stand out more. Before such a thresholding operation, SOBEL had provided a more noisy edge detection map than TIN. However, after thresholding we observe that while some edges detected by TIN were preserved, some edges detected by the SOBEL method where removed. As TIN preserved better significant edges, we proceed for further analysis with the edges detected by the TIN method.

\begin{figure}[!h]
  \vspace{-0.2cm}
  \centering
  \includegraphics[width=0.6\linewidth]{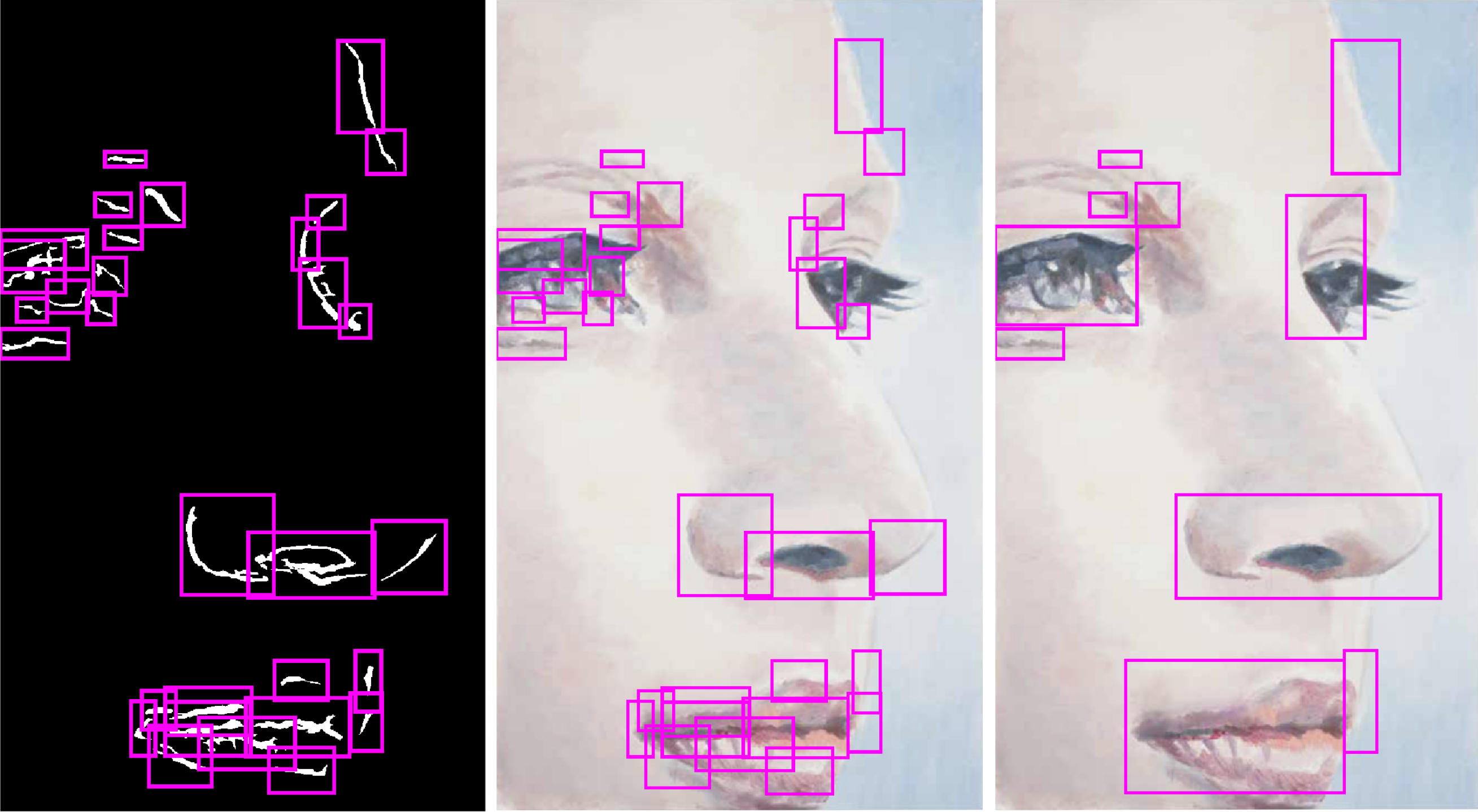}
  \caption{Computing the centres of interest based on edge difference maps. (i) 
Aggregation of edge differences. (ii) Projection of aggregation on painting. (iii) 
Expansion of bounding boxes.}
  \label{fig:interest-centres}
  \vspace{-0.1cm}
\end{figure}

We now can compute centres of interest, being those areas on the painting for which 
there are significant differences between the two edge maps. The algorithm proceeds in three  
steps: [i] Compute first the bounding boxes around the thresholded edges {\it only} contained in the painting, so neither in the picture only nor in both. Only those edges are retained which form an area 
containing more than $a$ points (pixels) with a perimeter containing more than $p$ points. The bounding
box is extended slightly (with an addition of $c$). [ii] Concate neighbors of bounding boxes by merging overlapping boxes from step [i] into larger boxes. 
[iii] Step [ii] is iterated until there are no overlapping bounding boxes. 
The different parameters $a$, $p$, and $c$ allow us to tune these steps to get more or fewer 
centres of interest. Results are shown in Figure \ref{fig:interest-centres}.  (i) shows the 
result of step [i] with $a = 100$, $p = 70$, and $c = 0.0023$. (ii) shows the result of step [i] projected on the painting. (iii) shows the result of applying step [ii] for two iterations at which point there are no more overlapping bounding boxes. 

We see in Figure \ref{fig:interest-centres},(iii) that the following 
centres are proposed: (a) the lips area at the bottom, specifically 
the right side of the lips ('right' from the perspective of the person 
being depicted), (b) the right eye and the area around it, specifically the region where
the eyebrow reaches the top of the nose, the pupil and the line under the 
right eye, (c) the nose area with the nostril and the curve at the right wing of the nose, 
(d) the left eye, particularly the corner with the nose, and (e) the region above the left eye.

\section{STEP IV. Deconstructing possible meanings} 
\begin{figure}[!h]
  \vspace{-0.2cm}
  \centering
  \includegraphics[width=0.5\linewidth]{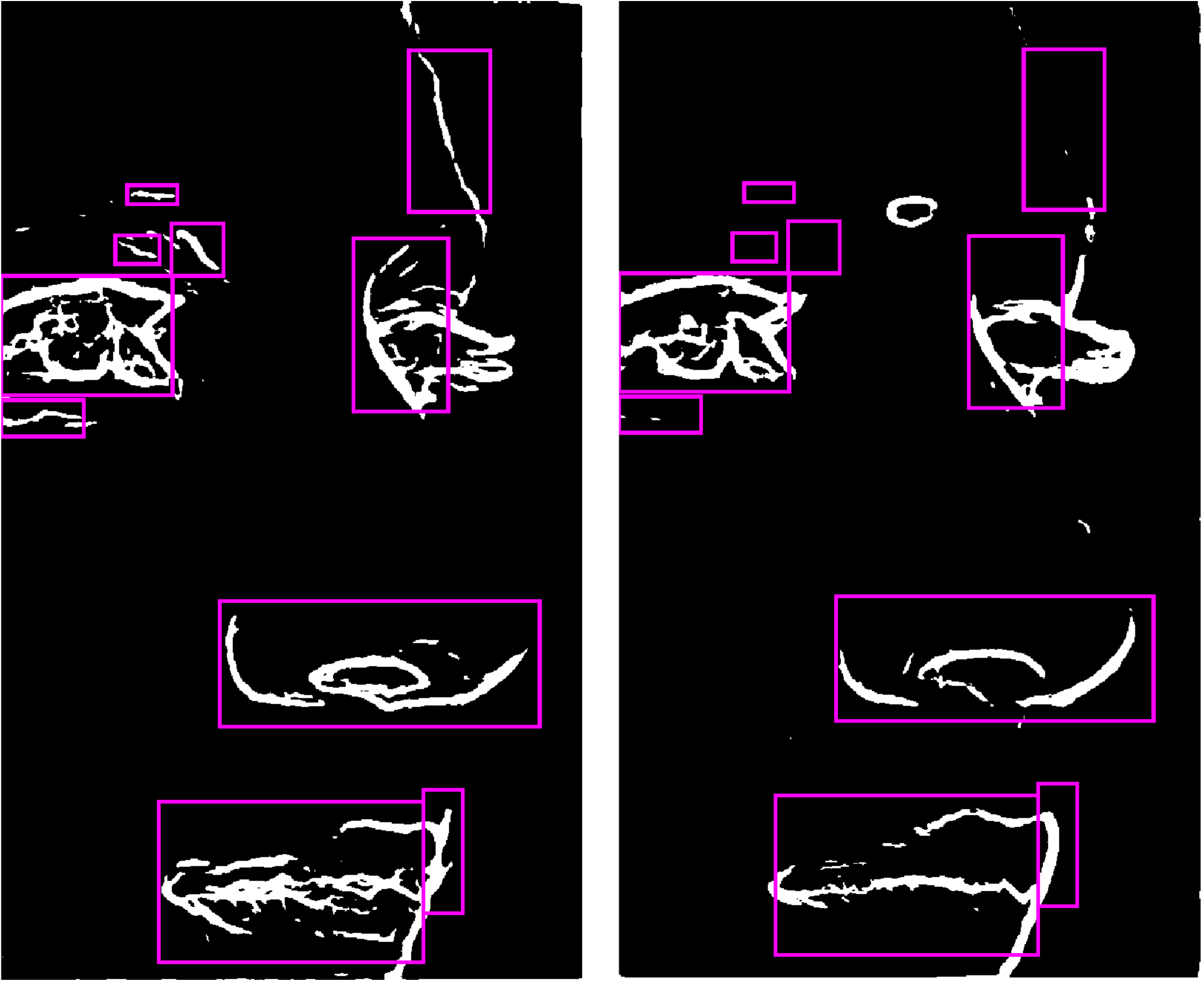}
  \caption{Edge maps with centres of intrest (TIN method). (i) Projection on edge map of
painting, (ii) Projection on edge map of original photograph.}
  \label{fig:edgemaps}
  \vspace{-0.1cm}
\end{figure}

Do these newly discovered centres of interest make any sense? A closer look at the original 
photograph and the painting shows that there is actually a 
subtle difference in facial expression between the photograph and the painting. This 
becomes clearer if we compare the centres of interest projected on the edge map 
of the painting (ii) and the photograph (iii) (see Figure \ref{fig:edgemaps}).

We see that on the painting, the facial expression of K. is 
more open and presents a weak smile, whereas on the photograph, the expression is more sturdy and 
closed. The more open facial expression is achieved by very subtle changes. The left corner 
of the lips in the painting is curled 
upward, the nostril more pronounced, the eye pupil bigger, the eyebrows showing a clearer V-shape, 
and the line under the left eye being more visible. These changes are subtle and unconscious 
to the viewer but they have an impact on the interpretation process, as they all point to 
a more open weakly smiling facial expression.

To see whether this interpretation has any validity. It is instructive to consider a `gold standard'
narrative about painting K. offered by Marc Donnadieu, chief curator of 
the Mus\'{e}e de l'Elys\'{e}e in Lausanne and published 
in the guidebook of the La Pelle exhibition at the Palazzo Grassi in Venice:\cite{Donnadieu:2019} 
\begin{quotation}
\vspace{-0.2cm}
"The artist was inspired by advertising billboards he saw in Panama, which feature women's faces that have been smoothed out to the point of erasing their personality. As a filmmaker would do, he zoomed on this face so close that it is partially cut out and incomplete.
By moving his `paintbrush camera' so close 
he treats this female face as if it was an object, which is precisely the purpose of advertising, in particular when selling beauty products. This approach renders the face empty, almost dead, especially because it is not contextualized.
But by zooming this way, the artist also highlights the gaze of this woman who has been so objectified that she doesn't even have a name, just a letter, K. And her gaze is very expressive, as if she tried to exist beyond the image and the commercial profit that is sought through her. She seems defiant, aware that she is exploited and ready to stand up as she looks far ahead, maybe towards a future where women will not be treated as objects.
The treatment is smooth, flat, and the pastel colors highlight the contrast between the artificial aspect of advertising imagery and the humanity of all women. K's mouth is shut, but she smiles discreetly and her silence speaks volumes."
\end{quotation}
The painting may be inspired by billboards in Panama on beauty products, although 
we have seen that the direct inspiration are fashion models from an advertising campaign of Dior.
Nevertheless the way women are represented is undoubtly similar. There is an {\it objectification} 
\cite{Fredrickson:1997} of the human body, more concretely in this case of the human face. 
This objectification is present in the photography but even more so in the painting: 
The extreme focus on the face so that there is an almost complete elimination of the context, 
the use of pastel color, the flat treatment of the skin to smoothen out details that make natural
faces alive, the choice of a letter K. for the title of the painting, 
instead of a real name. Marc Donnadieu mentions expressional meanings like: `smile 
discreetly', `defiant', `expressive gaze' which 
confirm the subtle changes in lips, eyes, eyebrows and nose that the edge-based 
comparative methodology detected. This example therefore illustrates the idea that AI methods can act
like a microscope that draws attention to interest regions and helps us see details that 
would otherwise remain unconscious. 

\section{A second case study} 

\begin{figure}[!h]
  \vspace{-0.2cm}
  \centering
  \includegraphics[width=0.8\linewidth]{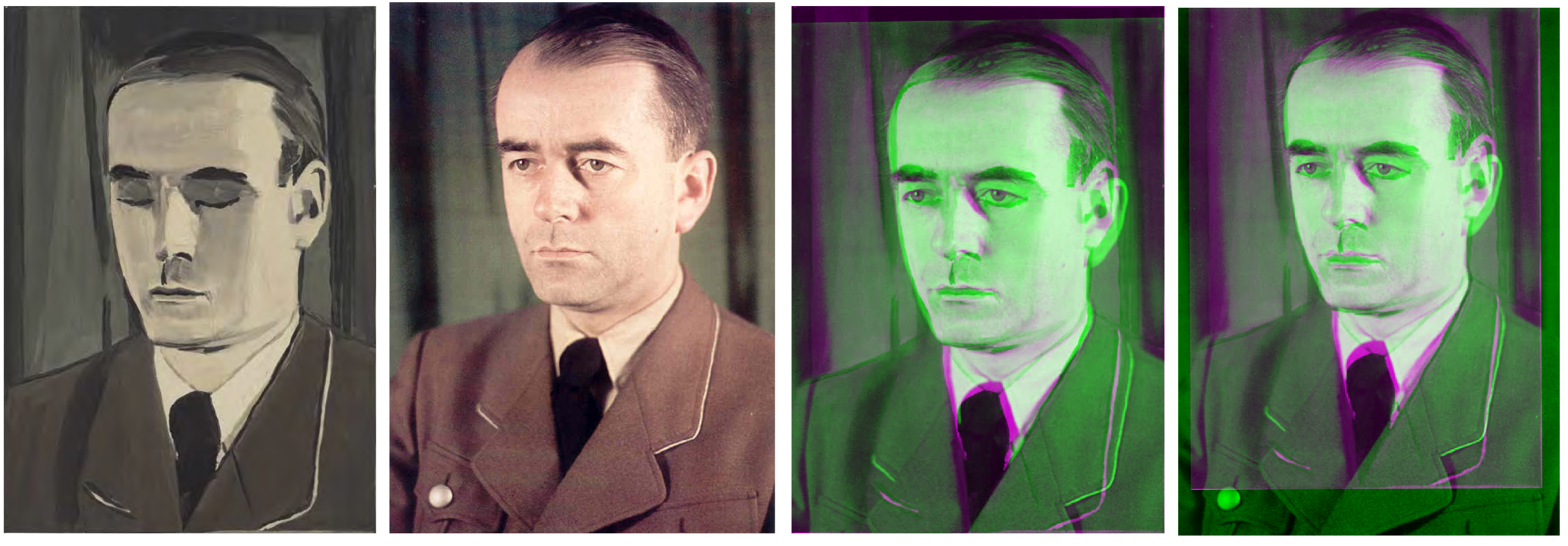}
  \caption{From left to right: (i) painting, (ii) original photograph, (iii) alignment 
picture on painting (iv) aligning of painting on picture.}
  \label{fig:secrets-painting-original-alignment}
  \vspace{-0.1cm}
\end{figure}

We now show very briefly a second example from the 2019-2020 Palazzo Grassi solo
exhibition by Luc Tuymans, a painting entitle Secrets. It depicts one of the main figures of 
the Nazi regime, Albert Speer, who has always denied any knowledge of the Holocaust atrocities. 
Figure \ref{fig:secrets-painting-original-alignment} shows first the painting as well 
as the source image, which has been cropped from a larger image that shows Speer in full Nazi regalia. 
Figure \ref{fig:secrets-painting-original-alignment} shows next the results 
of using the same alignment process as used for K., 
both for aligning the picture on the painting and the painting on the picture.

\begin{figure}[!h]
  \vspace{-0.2cm}
  \centering
  \includegraphics[width=0.7\linewidth]{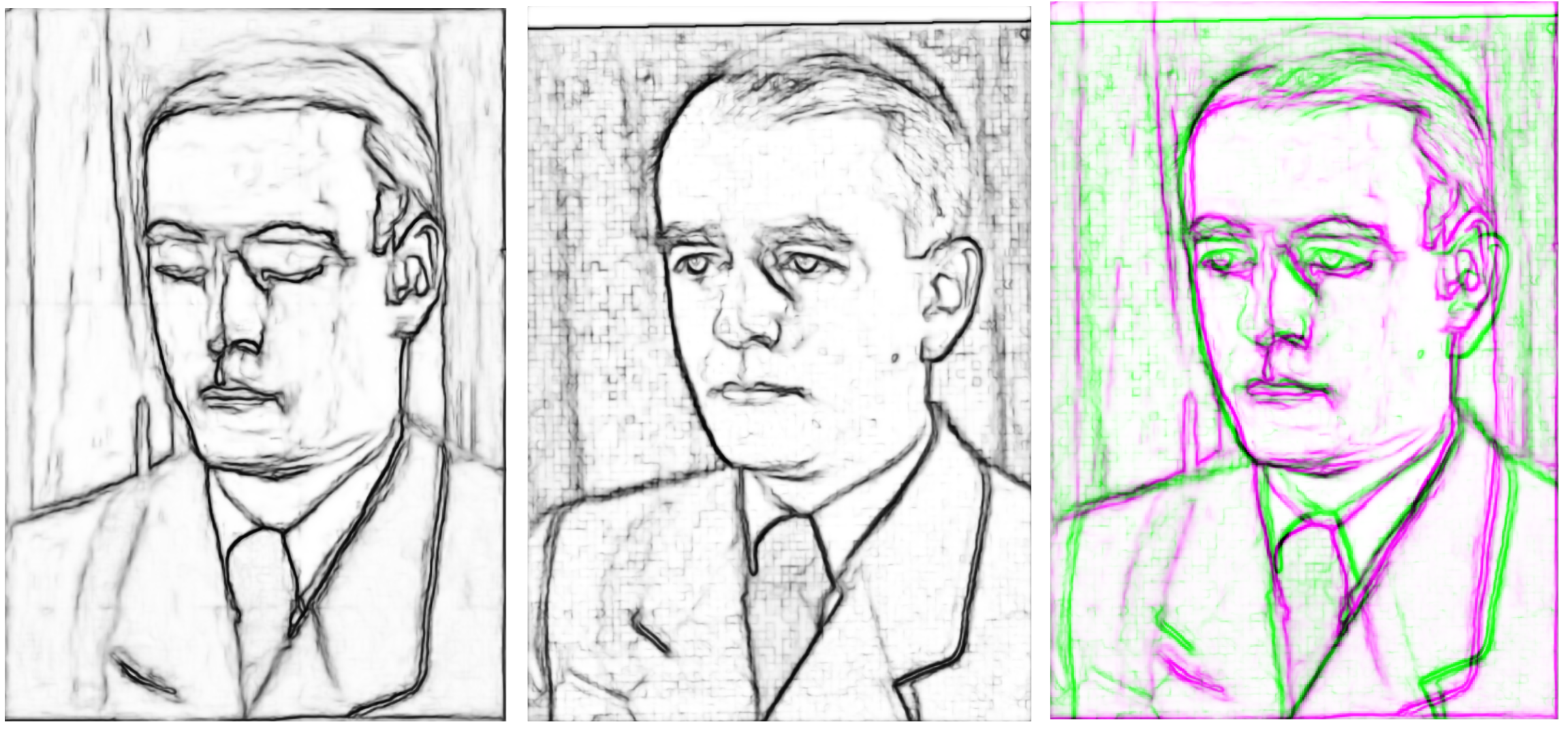}
  \caption{From left to right: (i) edges from painting, (ii) edges from aligned photograph, 
(iii) edge difference map painter's perspective.}
  \label{fig:results-edge}
  \vspace{-0.1cm}
\end{figure}

\begin{figure}[!h]
  \vspace{-0.2cm}
  \centering
  \includegraphics[width=0.8\linewidth]{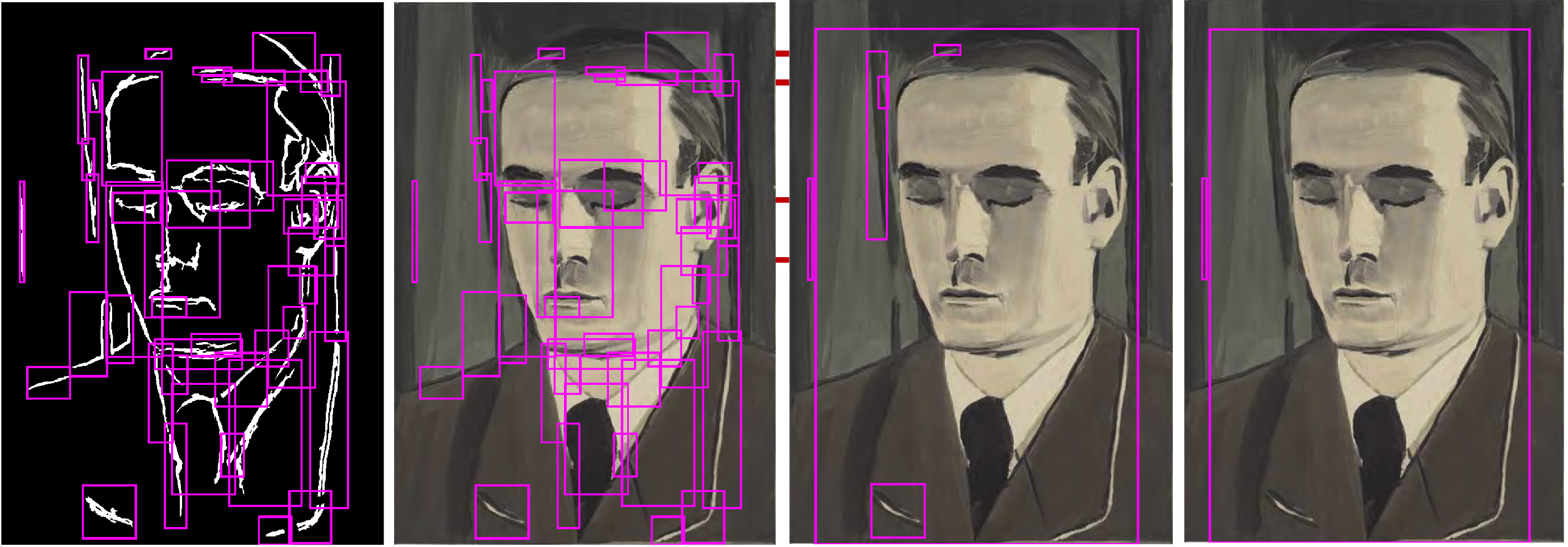}
  \caption{Computing the centres of interest based on edge difference maps. (i) 
Aggregation of edge differences. (ii) Projection of aggregation on painting. (iii) 
Expansion of bounding boxes.}
  \label{fig:centres-of-interest-secrets}
  \vspace{-0.1cm}
\end{figure}

We see that the alignment algorithm works very well, even though this case is much more 
challenging as the face is slightly rotated, stretched along the y-axis and squeezed along 
the x-axis. 
\noindent Figure \ref{fig:results-edge} shows the results of edge detection using the TIN method
with the difference map projected on the photograph. Again the results are very satisfactory. 
Finally, Figure \ref{fig:centres-of-interest-secrets} shows the same operations 
as in Figure \ref{fig:interest-centres}, now on the edge difference map of `Secrets'. 
Again we have to ask whether the regions identified through this process function as meaningful
centres of interest. Unfortunately there are clearly too many regions to allow a 
meaningful analysis, mainly because there 
are a lot of changes that the artist has made to the original source image. These changes 
are all purposeful. They have to do with making the face more geometric, 
closed-off, inward-looking and in denial. Most probably using other aggregation algorithms 
could lead to more useful results. 
\section{Conclusions} 

It is well known that painters never attempt to paint exactly a chosen image (or a real world scene)
but transform it to achieve an artistic purpose - except 
in artistic movements where realism is in itself an artistic statement, like in hyper-realistic art. 
This paper reported on attempts to deconstruct algorithmically what 
transformations a contemporary painter carried out and discussed in how far paying attention 
to the regions identified by this process helps us in the construction of a narrative capturing
the meanings invoked by a painting. AI algorithms are used here as 
microscopes that allow us to look in detail at transformations which are normally only experienced 
at a subconscious level. 

This paper focused on edge differences only. In one case study (the painting entitled K.), the regions 
found through comparing edge differences indeed point to centres of interest that can be 
interpreted. In the second case study (`Secrets') alignment and edge detection worked well but 
useful conclusions in terms of centres of interest could not be reached. 
These examples show that the comparative method is only at its first beginning. There are other 
dimensions of transformation that must be studied in equal detail and 
color is the most likely candidate. More generally, the hard but certainly fascinating, work
remains to see how far the type of analysis proposed 
here can help to recognize and label the objects in a painting, detect the general mood, the 
emotions being expressed by the depicted figures, and much more, all in the service of 
supporting narrative construction in viewers. 
%
%
%
%

\end{document}